\documentclass[conference]{IEEEtran}
\IEEEoverridecommandlockouts
\usepackage{cite}
\usepackage{amsmath,amssymb,amsfonts}
\usepackage{algorithmic}
\usepackage{graphicx}
\usepackage{textcomp}
\usepackage{xcolor}
\def\BibTeX{{\rm B\kern-.05em{\sc i\kern-.025em b}\kern-.08em
    T\kern-.1667em\lower.7ex\hbox{E}\kern-.125emX}}

\usepackage[symbol]{footmisc}
\usepackage{subcaption}
\usepackage{hyperref}
\usepackage{amssymb}
\usepackage{booktabs}
\usepackage{siunitx}
\usepackage[font=footnotesize]{caption}

\newcommand{\tocite}[1]{\textcolor{red}{[CITE]}}
\newcommand{\todo}[1]{\textcolor{orange}{[TODO]}}
\newcommand{\tocheck}[1]{\textcolor{cyan}{[CHECK]}}

\begin{document}



\title{Using Fitts' Law to Benchmark Assisted Human-Robot Performance\\}


\author{Jiahe Pan$^{1}$, Jonathan Eden$^{1}$, Denny Oetomo$^{1}$, and Wafa Johal$^{1}$
\thanks{
This work was partially supported by the Australian Research Council (Grant Nos. DE210100858 and DP240100938). Thanks to Audrey Balaska for providing feedback on this manuscript.}
\thanks{$^{1}$The authors are with the Faculty of Engineering and Information Technology, The University of Melbourne, Australia.
{\tt\footnotesize jmpan@student.unimelb.edu.au, \{eden.j, doetomo, wafa.johal\}@unimelb.edu.au}}%
\thanks{Digital Object Identifier (DOI): see top of this page.}
}

\maketitle

\begin{abstract}
Shared control systems aim to combine human and robot abilities to improve task performance. 
However, achieving optimal performance requires that the robot's level of assistance adjusts the operator's cognitive workload in response to the task difficulty. Understanding and dynamically adjusting this balance is crucial to maximizing efficiency and user satisfaction.
In this paper, we propose a novel benchmarking method for shared control systems based on Fitts' Law to formally parameterize the difficulty level of a target-reaching task. With this we systematically quantify and model the effect of task difficulty (i.e. size and distance of target) and robot autonomy on task performance and operators' cognitive load and trust levels. Our empirical results \textit{(N=24)} not only show that both \textit{task difficulty} and \textit{robot autonomy} influence task performance, but also that the performance can be modelled using these parameters, which may allow for the generalization of this relationship across more diverse setups. 
We also found that the users' perceived cognitive load and trust were influenced by these factorsw Given the challenges in directly measuring cognitive load in real-time, our adapted Fitts' model presents a potential alternative approach to estimate cognitive load through determining the difficulty level of the task, with the assumption that greater task difficulty results in higher cognitive load levels.s
We hope that these insights and our proposed framework inspire future works to further investigate the generalizability of the method, ultimately enabling the benchmarking and systematic assessment of shared control quality and user impact, which will aid in the development of more effective and adaptable systems.
\end{abstract}




\section{Introduction} \label{intro}

Teleoperation is a common form of Human-Robot Collaboration (HRC) used in contexts where autonomous operation is not feasible and human guidance is essential \cite{darvish2023teleoperation}. From satellite maintenance to medical surgery, teleoperation tasks often require efficient execution of precise movements between multiple targets using the robot arm's end-effector. 
As a result, operators are often burdened with high cognitive loads \cite{moniruzzaman2022teleoperation} when completing these tasks. To address this, shared control systems have been developed \cite{lin2020shared, jain2019probabilistic, young2019formalized} that leverage the complementary strengths of humans and robots to enhance task performance and reduce user workload.
Here, a key factor is the task difficulty (i.e. how challenging the task is), which often relates to characteristics of the teleoperation task such as the required accuracy and the time constraints. 
Previous works have shown that task difficulty can significantly impact user performance and cognitive load during HRC \cite{luo2021workload, melnicuk2021effect, bequette2020physical}, such that they should be considered when designing adaptive shared control systems. 




One important consideration when designing adaptive shared control is in assessing performance relative to task difficulty. Existing evaluations often employ tasks which are loosely defined and lack formal quantification of their difficulty, which hampers the replicability and generalizability of their research findings \cite{leichtmann2022crisis}. To the best of our knowledge, there is currently no common benchmark for evaluating the efficacy and efficiency of robot assistance in teleoperation, making it challenging to assess the true effectiveness of shared control systems in real-world HRC scenarios.

Fitts' Law \cite{fitts1954information} has been widely used in Human-Computer Interaction (HCI) to model the relationship between difficulty of reaching motions and human performance, typically measured through \textit{movement time}, which is the time it takes to complete the reaching motion. This model has served as a foundational framework in HCI enabling the generalization of research findings and the optimization of user interfaces. 
Its widespread adoption as a benchmarking tool has significantly contributed to the understanding and design of effective user interfaces \cite{argelaguet2013survey, drewes2010only, so2000effects}.

Applying a similar benchmarking framework to HRC presents unique challenges. Specifically, the assumptions underlying Fitts' Law regarding human behavior may not hold when robot assistance is introduced, as the dynamics of shared control alter the interaction landscape. It is thus essential to investigate the influence of robot assistance on the relationship between task difficulty and human performance.
To employ Fitts' Law as an effective benchmarking tool in HRC, we must first understand how various levels of robot assistance affect the difficulty-performance relationship it predicts. Gaining such insights would not only validate the applicability of Fitts' Law in HRC contexts but could also inform the development of adaptive control strategies. For instance, understanding these dynamics could enable the real-time adjustment of robot assistance levels based on predicted human performance, thereby enhancing the effectiveness and safety of shared control systems.

Motivated by this, the overarching research question we aim to address is: \textit{How does task difficulty and robot assistance affect the user's performance and cognitive load?}
To better understand this relationship, we conducted a teleoperation study using a target reaching task with a definition of difficulty based on Fitts' Law (See Section \ref{method:task}), and a shared-control scheme which allowed continuous variation of the robot assistance level\footnote{The code and data, along with a demo experiment video, are available on the project website: \href{https://sites.google.com/view/autonomyfitts/home}{https://sites.google.com/view/autonomyfitts/home}.}. Target reaching was chosen to also reflect real-world manipulation tasks which are often composed of such atomic motions. Through both objective and subjective measures of task performance and cognitive workload, and the use of Fitts' Law as a more formal and generalizable evaluation of the effects of robot assistance on our measures under systematic variation of task difficulty, we aimed to uncover the impact of task difficulty and level of robot assistance on these variables.

To summarize, the paper contributes the following:
\begin{enumerate}
    \item It provides empirical evidence of the effects of task difficulty and assistance on performance and cognitive load in shared control.
    \item It introduces an adapted Fitts' Law as a benchmarking method for evaluating shared control systems for reaching tasks. This captures the interaction between task difficulty and robot assistance on user performance.
\end{enumerate}

\section{Related Works} \label{related}

\subsection{Effects of Shared Control on Performance and Perception} \label{related:shared_control}
By using human intelligence to drive robots via a human-machine interfaces \cite{cui2003review}, teleoperation has become a robust method for robots to assist humans to perform real-world tasks in uncertain and unsafe environments \cite{alvarez2001reference}. To enable the combining of the respective strengths of the robot and the human for successful complex task completion, teleoperation often uses shared-control \cite{kent2017comparison}, where command inputs from the human and an autonomous controller are arbitrated to determine the robot's resulting actions. A variety of formulations exist for implementing shared control, including game-theoretical formulations of the human-robot system \cite{music2017control} and control-blending mechanisms \cite{dragan2013policy}. Existing shared-controllers have been designed to optimize for task performance, such as by minimizing collisions with obstacles while navigating \cite{deng2019bayesian} or by maximizing success rate in object grasping \cite{zhuang2019shared}. Recent works have also shown that the human operator's internal state including both cognitive load and their trust perception of the robot can impact performance and interaction quality \cite{pan2024effects, hopko2021effect, kok2020trust}, thus highlighting them as importance factors to consider when designing shared control.

However, to design an effective shared-controller which adapts to the user's internal state, it is crucial to have reliable real-time measures such as for cognitive load. Cognitive load is an indicator of task complexity based on the number of conceptual elements that need to be held in the user's mind, at any one time, to solve a specific task \cite{plass2010cognitive}. Subjective assessment schemes such as the widely-employed Task Load Index (NASA-TLX) \cite{NASA-TLX} have been shown to reliably capture cognitive load perception \cite{marchand2021measuring} but do not allow for real-time measurements that could be used in adaptive shared-control. Other methods such as the dual-task paradigm of performing a secondary task (e.g., reproducing melodies \cite{park2015rhythm} or performing simple math \cite{lee2015influence}) concurrently with a primary task, and physiological measures (e.g., pupil dilation \cite{eckstein2017beyond} or brain activity \cite{miller2011novel}) allow objectively measuring the cognitive load level at higher frequencies \cite{marchand2021measuring} but suffer from the impact of confounding factors such as learning effects \cite{esmaeili2021current} and expertise in multitasking \cite{strobach2015better}. 
Therefore, while it may be beneficial to employ multiple measures of cognitive load at once, identifying a reliable real-time method to capture and quantify cognitive load remains a challenging open problem. 

Similarly, adapting robot behavior based on the user's trust is challenging due to the lack of an efficient and reliable real-time trust measure.
Human trust in an agent has been defined as a multidimensional latent variable that mediates the relationship between events in the past and the human's choice of relying on the agent in an uncertain environment \cite{kok2020trust}. Trust is an internal measure experienced by humans, making it difficult to capture objectively \cite{chita2021can}. Existing measurement methods include physiological information \cite{kohn2021measurement} and brain imaging \cite{wang2018eeg} which can be influenced by confounding factors. As a result, subjective questionnaires such as the Multi-Dimensional Measure of Trust (MDMT) \cite{mdmt1} which captures trust across 2 categories - \{\textit{Capacity Trust, Moral Trust}\} are most widely employed in HRC \cite{ahmad2022no, ullman2021challenges}.

\subsection{Quantifying Difficulty and Performance using Fitts' Law} \label{related:fitts_law}
\begin{figure}[t]
\centering
\includegraphics[width=5cm]{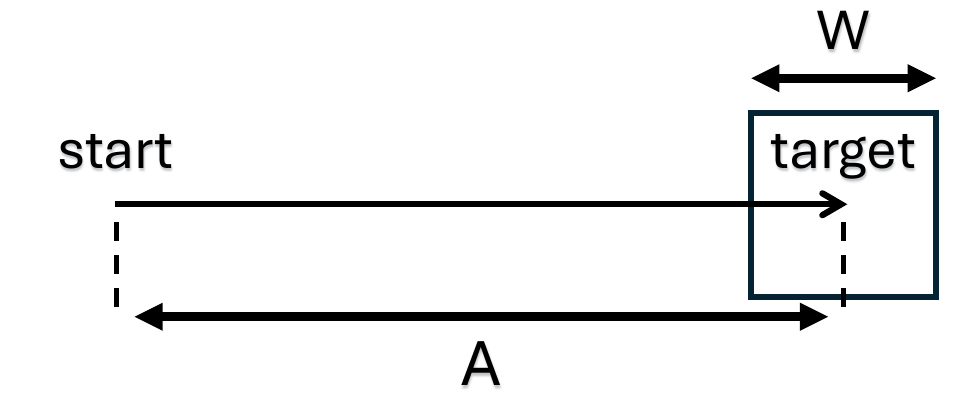}
\caption{An illustration of Fitts' Law. Amplitude (A) and Width (W) determine the index of difficulty of the reaching motion.}
\label{fig:fitts_1d}
\end{figure}

Fitts' Law \cite{fitts1954information} is a widely used human performance model that has been applied to evaluate interface design in HCI \cite{argelaguet2013survey, drewes2010only, so2000effects}. The original formulation predicts the \textit{movement time} (MT) to reach a target using an \textit{index of difficulty} (ID), where
\begin{equation}
\begin{aligned}
    \mathrm{MT} = a + b \cdot \mathrm{ID}, \quad \mathrm{ID} = \log_2\left(\frac{A}{W} + 1\right).
\end{aligned}
\label{eqn:fitts_og}
\end{equation}
Here, the model parameterizes reaching motions using the straight-line distance from the starting point to the target (\textit{amplitude}) $A$ and the target error margin (\textit{width}) $W$ (See Fig.\, \ref{fig:fitts_1d}). Using these two variables, the ID (measured in bits) can be related to the MT through \eqref{eqn:fitts_og}, where $a$ and $b$ are the calculated intercept and slope of this linear relation. 
Other works have also extended the original formulation to higher-dimensions \cite{wang1999object, kulik2020motor} and trajectory tracking \cite{accot1997beyond}.


Fitts' Law is chosen as a building block for HRC benchmarking it enables a systematic definition of task difficulty which can be varied continuously, thereby enhancing result generalizability. Furthermore, it establishes a method to evaluate movement performance in reaching motions through the empirically-validated linear relationship between \textit{movement time} and \textit{index of difficulty}, and therefore allows us to directly examine the impact of robot autonomy.

Here, we formulate the two main hypotheses of this study:
\begin{itemize}
    \item \textbf{H1} - Task performance will decrease with greater task difficulty and increase with more robot autonomy.
    \item \textbf{H2} - Cognitive load will increase with greater task difficulty and decrease with more robot autonomy.
\end{itemize}



\begin{figure}
\centering
\begin{subfigure}{0.24\textwidth}
    \includegraphics[width=\textwidth]{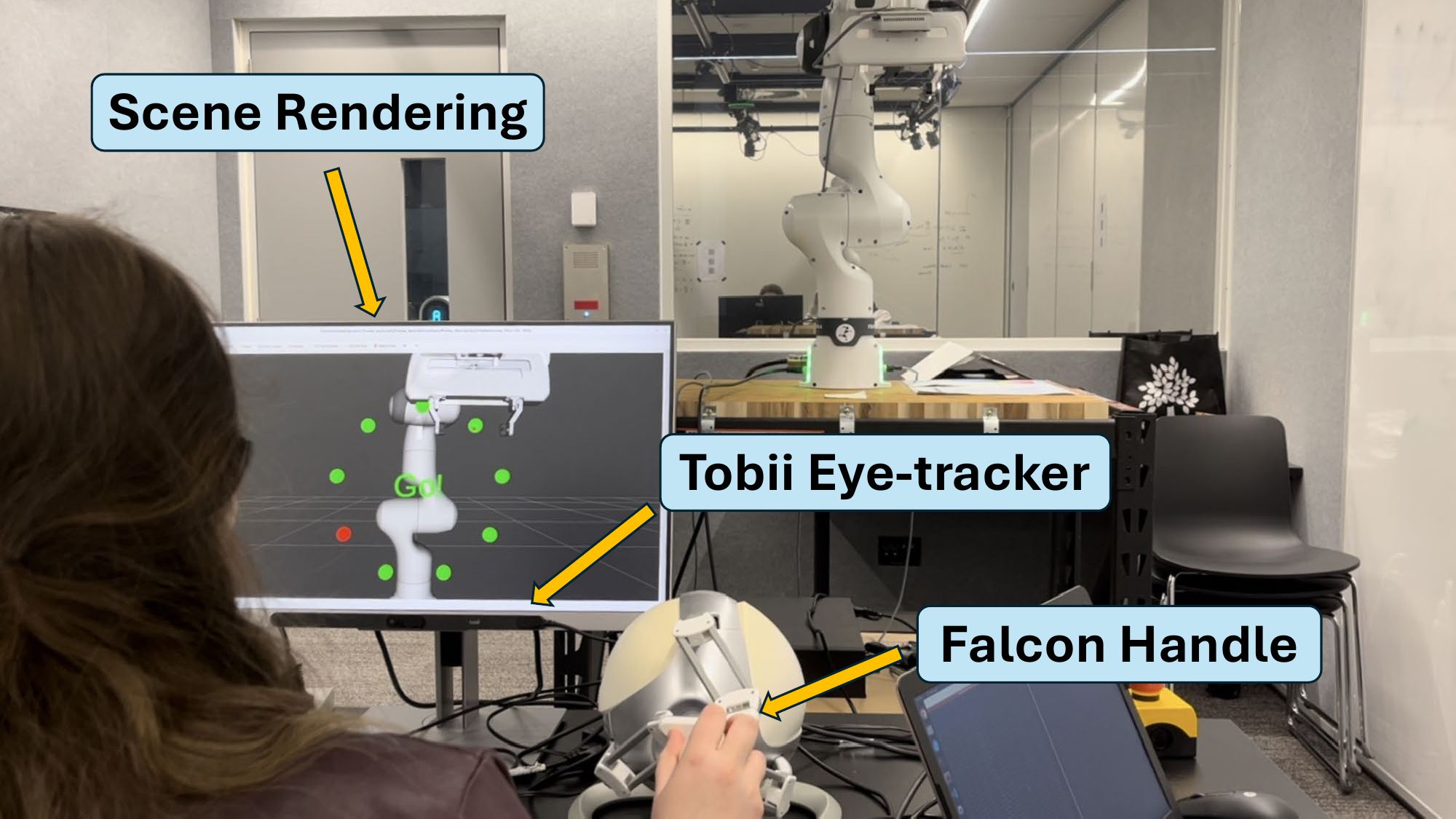}
    \caption{The target-reaching task was performed via a Novint Falcon controller and visual feedback on the screen.}
    \label{fig:setup}
\end{subfigure}
\begin{subfigure}{0.24\textwidth}
    \includegraphics[width=\textwidth]{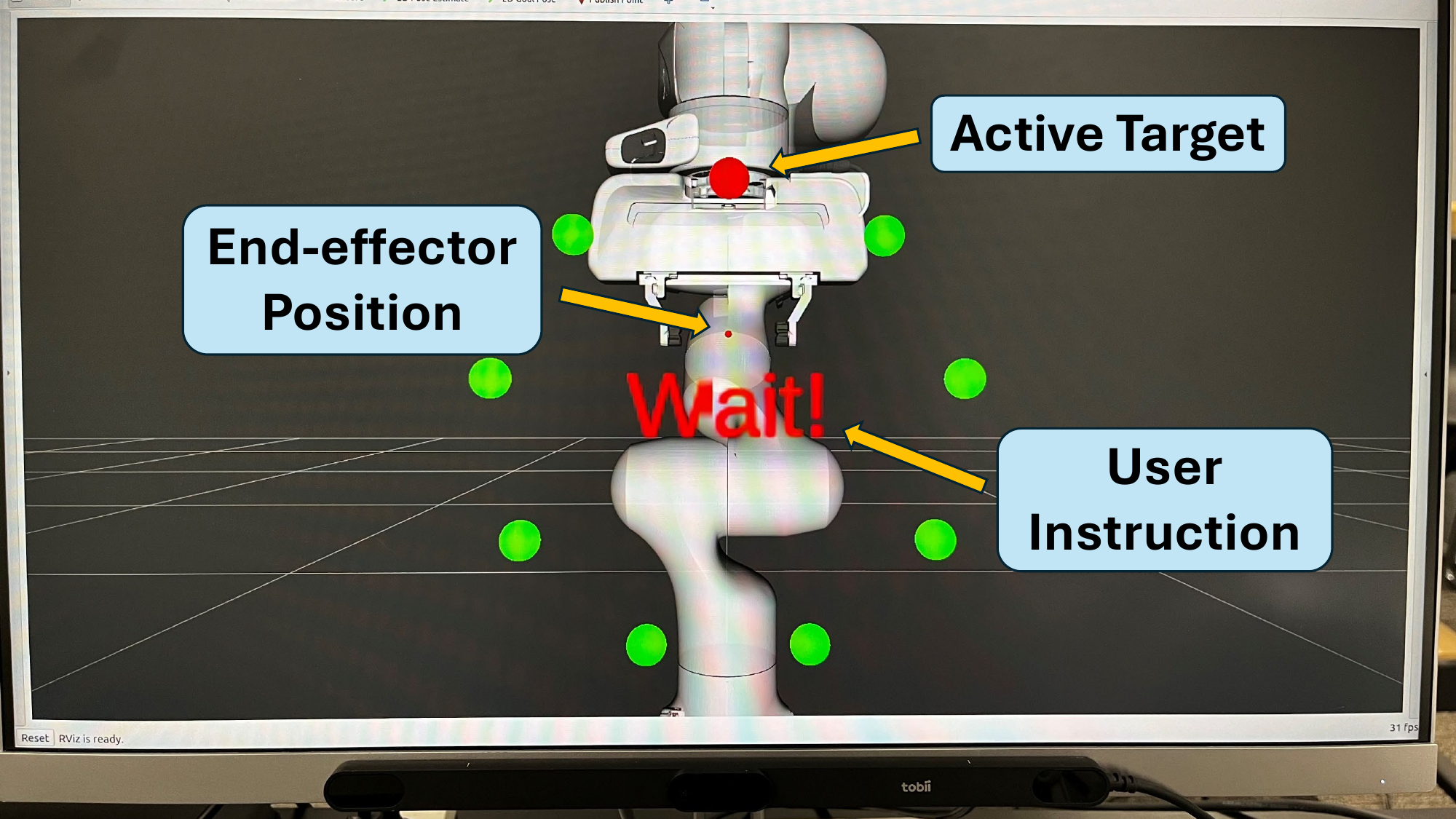}
    \caption{The ring of targets and the position of the center of the end-effector at each time instant were displayed.}
    \label{fig:rviz}
\end{subfigure}
\caption{Experimental setup (a) and the task visualization (b).}
\label{fig:entire_setup}
\end{figure}

\section{Method} \label{method}
We evaluated the performance, cognitive load and trust of participants during their teleoperation of a Franka Emika Research 3 robotic arm (Fig.\,\ref{fig:setup}). This employed a target-reaching task as in other Fitts' Law studies \cite{wagner2023fitts, brickler2020fitts}.

\subsection{User Interface} \label{method:user_interface}
The user interface was inspired by a typical teleoperation scenario, where the participant teleoperated the real robot via a Novint Falcon haptic interface, and perceived the physical robot through an RViz \cite{kam2015rviz} virtual rendering (Fig.\,\ref{fig:rviz}), which also displayed a ring of targets. The Falcon was connected to the robot via Ethernet to minimize communication delay. The ring of targets was designed to lie in a vertical plane with a constant depth, where the Falcon's depth was fixed via a high-gain PID controller with a zero depth reference. No haptic feedback was provided to participants. Under this setup, participants were able to freely move the Falcon within the vertical plane without any disturbances in the depth direction.

\begin{figure}
\centering
\begin{subfigure}{0.2\textwidth}
    \includegraphics[width=\textwidth]{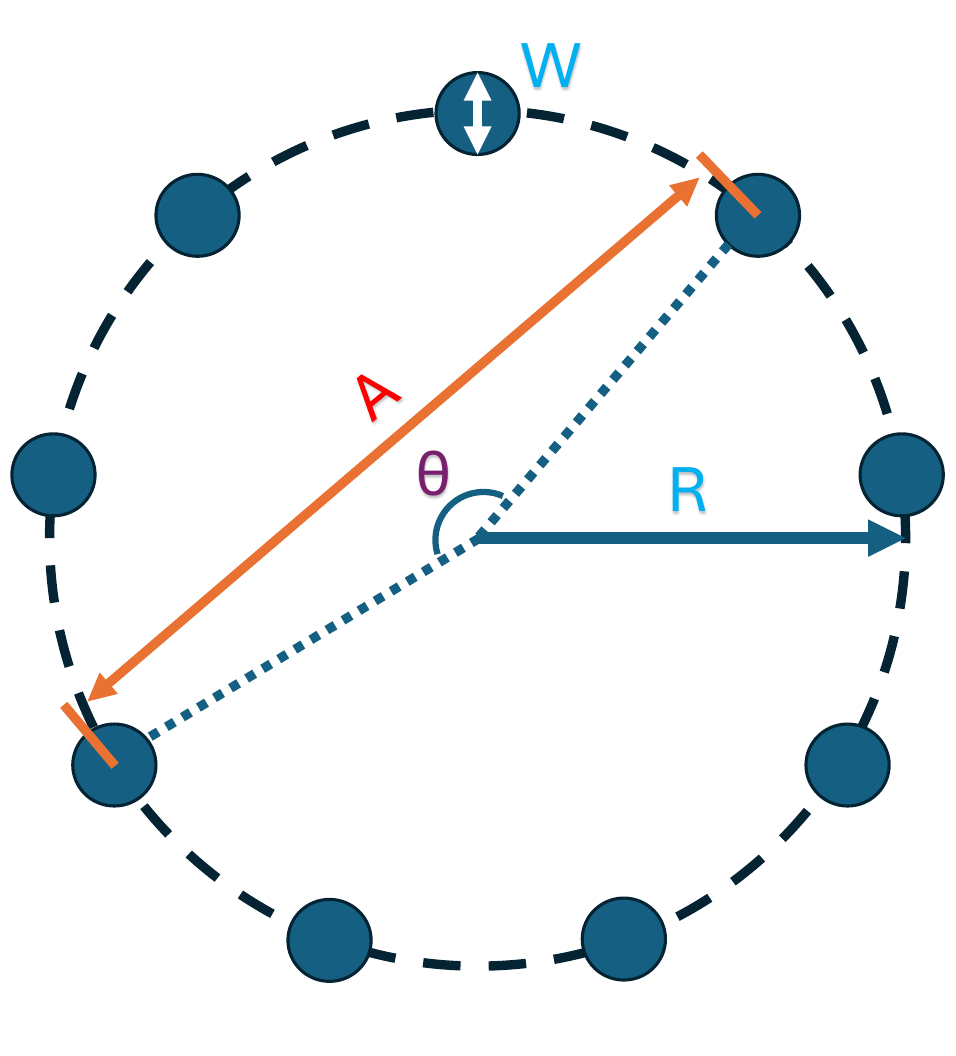}
    \caption{The ring radius (R), movement amplitude (A), target width (W), and angle between consecutive targets around the ring (\(\theta\)).}
    \label{fig:ring_geometry}
\end{subfigure}
\hspace{0.5cm}
\begin{subfigure}{0.2\textwidth}
    \includegraphics[width=\textwidth]{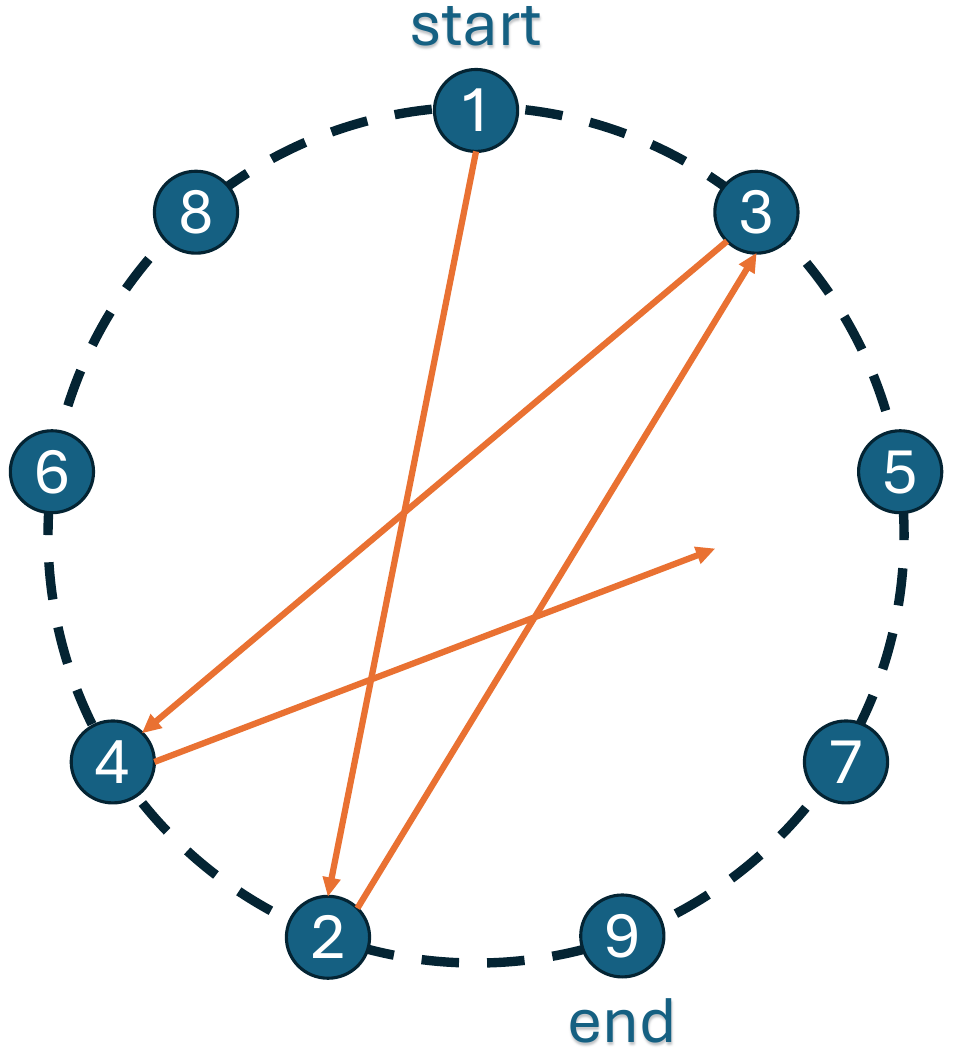}
    \caption{The reaching movement order for each ring of targets.}
    \vspace{0.55cm}
    \label{fig:reach_sequence}
\end{subfigure}
\caption{Visualization of the ring geometry (a) and reaching sequence (b).}
\label{fig:ring_info}
\end{figure}

\subsection{Task} \label{method:task}
The task entailed reaching for a ring of 9 virtual targets with the robot's end-effector in a pre-defined sequence (Fig.\,\ref{fig:reach_sequence}). The choice of this reaching task was motivated by several reasons. 
First, it has previously been used to benchmark robot abilities in real-world robot tasks \cite{suarez2020benchmarks, funk2021benchmarking} and has also been used to evaluate human movement and interface design in HCI \cite{wagner2023fitts, brickler2020fitts, monteiro2023evaluation}. This resonance with HCI user interface design may enable the transfer of existing HCI findings to better inform the design of intuitive and efficient user interfaces for robot teleoperation.
Moreover, reaching motions are a critical component common to most complex real-world tasks such as in surgical and industrial settings. 
Finally, the task enabled a clear definition of an \textit{index of difficulty} (ID) based on Fitts' Law, and allowed systematic variation of robot autonomy through a blending control scheme (see Section \ref{method:robot_autonomy}).

The reaching sequence was fixed across all trials (Fig.\,\ref{fig:reach_sequence}). Participants were asked to complete the task ``as quickly as possible''. The ring of circular targets was parameterized by two variables - the ring radius \(R\) and the target diameter \(d\). Successfully reaching a target requires the position of the robot's end-effector to be inside the circular target. The Fitts' parameters of each individual reaching motion between consecutive targets in the sequence was computed as
\begin{equation}
    A = 2 \hspace{0.03cm} R \sin{\frac{\theta}{2}}, \quad  W = d.
\label{eqn:reach_params}
\end{equation}
Here, \(\theta = 160^{\circ}\) was the angle between each pair of consecutive targets in the ring sequence. Using \(\theta\) and \(R\), the chord length between consecutive targets was calculated as the movement amplitude \(A\), while the movement width \(W\) was set to the target's diameter (see Fig.\,\ref{fig:ring_geometry}). The Fitts' ID for each reaching motion was then computed using \eqref{eqn:fitts_og}. We generated a set of four target rings with Fitts' ID summarized in Table \ref{table:ring_params}. Note, the center of the ring is positioned at a constant offset from the robot's base, which corresponded to the mapped position of the handle of the Falcon interface in its origin position. 
\begin{table}[t]
\centering
\begin{tabular}{l|lllll}
\hline
\textbf{Ring No.} & \textbf{R (m)} & \textbf{A (m)} & \textbf{W (m)} & \textbf{ID} & \textbf{RT (s)} \\ \hline
1                 & 0.06           & 0.118          & 0.02           & 2.788       & 0.6                     \\
2                 & 0.06           & 0.118          & 0.01           & 3.680       & 0.6                     \\
3                 & 0.12           & 0.236          & 0.02           & 3.680       & 1.2                     \\
4                 & 0.12           & 0.236          & 0.01           & 4.623       & 1.2                     \\ \hline
\end{tabular}
\caption{Ring geometry and the associated \textit{index of difficulty} (ID) for individual reaching motions. R, A and W represent the real-world ring radius, movement amplitude between consecutive targets, and the target width, respectively. RT is the set time for an autonomous robot reaching motion.}
\label{table:ring_params}
\end{table}

\subsection{Shared Control} \label{method:robot_autonomy}
The robot controller was designed to accurately follow an interpolated straight-line trajectory between consecutive targets with a constant velocity. The controller's reference completion time for each trajectory (see Table \ref{table:ring_params}) was set to be faster than the minimum recorded completion time of the same motion by a human expert. This ensured that for an average user, the robot provided reaching assistance rather than hindrance. Each reach started with the robot's end-effector at the center of the first target (Target 1 in Fig.\,\ref{fig:reach_sequence}). Participants were instructed to start each reaching motion as soon as the target activates, and the robot was programmed to do the same.

The robot was commanded through a blending control between the human and its own autonomous controller \cite{dragan2013policy,marcano2020review}, where the input position was computed as
\begin{equation}
\begin{aligned}
    \mathbf{u} = \gamma \mathbf{u_r} + (1-\gamma) \mathbf{u_h}, \quad
    \gamma &\in [0, 1] \subset \mathbb{R}.
\end{aligned}
\label{eqn:std_convex}
\end{equation}
Here \(\mathbf{u_r} \in \mathbb{R}^3\) denotes the reference end-effector position input generated by its autonomous controller, and \(\mathbf{u_h} \in \mathbb{R}^3\) denotes the reference end-effector position from the human's input which was read at 500\,Hz. The human input \(\mathbf{u_h}\) was set to be a 3 times scaling of the Falcon handle position, where this factor was determined using the ratio between the maximum ring diameter and the size of the Falcon's effective workspace.  The scalar \(\gamma\) represents the level of \textit{robot autonomy}, where \(\gamma = 1\) corresponds to complete robot control (\textit{full autonomy}) and \(\gamma = 0\) corresponds to complete human control (\textit{no autonomy}). By altering \(\gamma\) the relative autonomy between the human and robot was varied. This variable also determined the proportion of each reach that the robot will complete before stopping, under no user input. The resulting input \(\mathbf{u} \in \mathbb{R}^3\) was passed in ROS to the Kinematics and Dynamics Library (Orocos) inverse-kinematics solver to compute joint positions, which were tracked by the robot joint controllers running at 500\,Hz.

\subsection{Measures} \label{method:measures}
We used both objective and subjective measures to investigate the relationship between \textit{task difficulty} and the robot's \textit{autonomy level} in the target reaching task and the participants' task performance, cognitive load and trust.

\subsubsection{Demographics Information} \label{measures:demo_info}
We collected participant demographic information via an initial questionnaire. For their potential relevance in teleoperation, we included the following three items which participants rated on a 7-point Likert scale from 1 (\textit{strongly disagree}) to 7 (\textit{strongly agree}): i) \textit{``I trust new technology in general"}; ii) \textit{``I play computer games regularly"}; and iii) \textit{``I am proficient in a musical instrument"}.

\subsubsection{Performance Measure} \label{measures:in_task_measures}
In each trial, we recorded the autonomous controller inputs, user inputs, and the resulting end-effector movements from the combined inputs at \(40\)\,Hz. The task performance was then evaluated for each reach by computing the \textit{movement time (MT)} --- the time interval between successive target reaches using the end-effector. Separately recording the autonomous controller and user inputs was aimed to enable a more thorough understanding of the movement and interaction.

\subsubsection{Pupil Diameter} 
Larger pupil dilation has been associated with more intense cognitive processing \cite{eckstein2017beyond, white2017usability}. Therefore, we recorded pupil diameter data during each trial at 60 Hz using the Tobii Pro Spark screen-based eye-tracker. 

\subsubsection{Self-Reported Measures} \label{measures:questionnaire_measures}
After each ring, participants reported on the following measures via questionnaires:

\textbf{Perceived Autonomy} - The ``perceived autonomy questionnaire'' \cite{harbers2017perceived} has been employed in HRI research \cite{roesler2022influence, balatti2020method} to capture user perception of autonomy. Adapting the questionnaire, we included the 10-point discrete scale item: \textit{``How autonomous did you feel the robot was?''}. 

\textbf{Cognitive Load} - We administered the NASA-TLX questionnaire \cite{NASA-TLX} with all six original sub-scales in a randomized order as an assessment of cognitive workload.

\textbf{Trust} - The MDMT questionnaire \cite{malle2021multidimensional} is composed of 8 sub-scales in both \textit{Capacity} and \textit{Moral Trust}. Since our study focused on the human's trust of the robot's target reaching behavior, we chose to only include the 8 \textit{Capacity Trust} items.


\section{Study Design} \label{method:study_design}

\subsection{Conditions} \label{design:conditions}
To test our hypotheses, we used a counterbalanced, within-subject design to examine the effects of task difficulty and robot autonomy while controlling for order. The within-subject conditions were the \textit{ring number} \(\in\) \{\textit{1, 2, 3, 4}\} which manipulated the Fitts' ID as presented in Table \ref{table:ring_params}, and the \textit{autonomy} level which was set to \textit{no} (\(\gamma=0\)), \textit{medium} (\(\gamma=0.4\)) or \textit{high} (\(\gamma=0.8\)) autonomy. This gave \(4 \hspace{0.1cm} (\textit{ring number}) \times 3 \hspace{0.1cm} (\textit{autonomy}) = 12\) conditions, which were counterbalanced using a Latin square design. With these conditions, we performed an a priori power analysis to determine our sample size using G*Power \cite{faul2007gpower}. For a repeated-measure ANOVA with 0.8 power, \(\alpha\) = 0.05 and a medium effect size of \(f\) = 0.25, the calculation resulted in a sample size of \(N = 24\). 

\subsection{Participants} \label{design:participants}
Ethical approval for the study was granted by The University of Melbourne's Human Ethics Committee under project ID 27750. The 24 participants were aged from 18 to 30 y.o. (\(M=23.0, \hspace{0.1cm} SD=2.7)\), 15 identified as female and 9 as male. All participants performed the target reaching task using their preferred dominant hand (23 right-handed). Participants received a \$20 gift voucher as compensation at the end of the approximately 50 minute experiment.

\begin{figure}[h]
    \centering
    \includegraphics[width=0.5\textwidth]{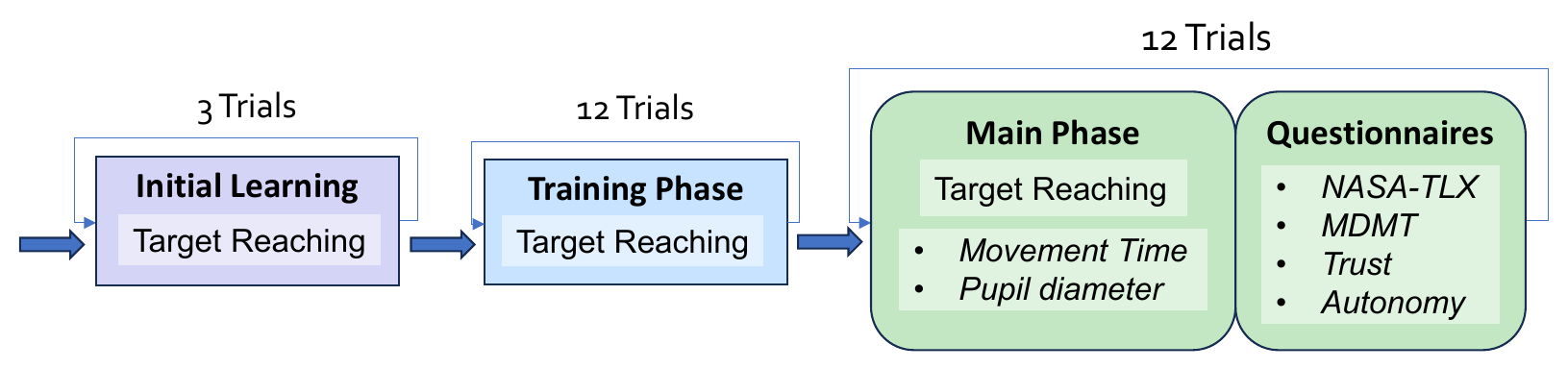}
    \caption{Experimental procedure including the training phase and main phase. Each trial is administered with a given ring of targets and autonomy level $\gamma$.}
    \label{fig:exp_flowchart}
\end{figure}

\subsection{Experimental Procedure} \label{method:exp_procedure}
Each participant took part in an experiment comprising of a \textit{training} and a \textit{main} phase (Fig.\,\ref{fig:exp_flowchart}). In each phase, participants completed all 12 conditions --- each of which was a single trial with a specific combination of ring number and robot autonomy level (see \ref{design:conditions}). In both phases, participants were informed that the robot's autonomy level might change between trials, but were not told the autonomy level itself nor any information on its ability to perform the task.

\subsubsection{Training Phase} \label{procedure:training_phase}
Participants were initially shown the setup and visualization (Fig.\,\ref{fig:setup}), followed by three practice trials under the \textit{no} (\(\gamma = 0\)), \textit{medium} \((\gamma = 0.4)\) and \textit{high} (\(\gamma = 0.8\)) autonomy levels, respectively, for them to familiarize themselves with the Falcon system, the setup and the reaching task. In these trials, participants were explicitly informed about the autonomy level, which they could use as a reference for perceiving the autonomy levels in the \textit{main} phase. They then completed 12 task trials matching the 12 respective conditions without knowledge of the robot autonomy levels. No data was recorded during \textit{training} as it was designed for the participants to practice the target reaching task with all combinations of target rings and robot autonomy levels, thereby minimizing the impact of learning effects on the \textit{main} phase.

\subsubsection{Main Phase} \label{procedure:main_phase}
In each \textit{main} phase trial, there was an initial 5\,s preparation window shown as a countdown on the screen (Fig.\,\ref{fig:rviz}), during which the target ring for that trial was also displayed. In addition, the robot's end-effector position was shown as a red dot located between its parallel grippers. After the 5\,s, the countdown text changed to the word \textit{``Go!"}, indicating that the trial has started. The active target was always colored red, with all other targets green. Upon a successful reach of the active target, it changed color to green, followed by the next target becoming red. Participants were instructed to ``as quickly as possible" complete the task, after which the on-screen text changed to \textit{``Stop!"}, indicating that the trial had finished. Each session took approximately 50 minutes, including the 10-minute introduction block and a 20-minute block for each of the \textit{training} and \textit{main} phases. During each trial, 8 movement times were recorded from the 8 intervals across the 9 targets and participants self-reported on the questionnaire after each trial. 

\section{Results} \label{results}
After all measures were verified to be normally distributed using the Shapiro-Wilk test, we performed two-way repeated-measures ANOVA with \textit{autonomy} and \textit{ring number} as within-subject variables. Here, using the \textit{ring number} as a single variable (instead of each ring's amplitude and width separately) enabled us to perform analysis in a style consistent with Fitts' Law and allowed us to validate whether rings 2 and 3 - which have the same Fitts' ID - would yield the same results as predicted by Fitts' Law. Results were considered significant at the threshold \(\alpha < .05\). For measures which yielded significant results, post-hoc pairwise comparisons with the Holm-Bonferroni correction were performed to further examine the effects of the within-subject variables on the measures. The ANOVA results are summarized in Table \ref{table:anova_results}.


\begin{table*}[t]
\centering
\begin{tabular}{c|cccc|cccc|cccc}
                   & \multicolumn{4}{c|}{\textit{\textbf{Ring number}}} & \multicolumn{4}{c|}{\textit{\textbf{Autonomy}}} & \multicolumn{4}{c}{\textit{\textbf{Interaction}}} \\ \hline
\textbf{Measure}   & DF     & F          & p        & $\eta^2$     & DF      & F         & p       & $\eta^2$    & DF     & F         & p         & $\eta^2$     \\ \hline
Perceived Autonomy & (3, 69)        & 3.530      & 0.019    & 0.021    & (2, 46)        & 61.731    & $<$0.001    & .472    & (6, 138)       & 1.428     & 0.208     & 0.013    \\
Movement Time      & (3, 69)        & 225.503    & $<$0.001     & 0.665    & (1.58, 36.42)   & 222.807   & $<$0.001    & 0.549   & (6, 138)       & 8.157     & $<$0.001      & 0.061    \\
NASA-TLX           & (3, 69)        & 6.859      & $<$0.001     & 0.040    & (1.24, 28.62)   & 15.271    & $<$0.001    & 0.082   & (6, 138)       & 3.823     & 0.001     & 0.021    \\
MDMT               & (3, 69)        & 8.069      & $<$0.001     & 0.025    & (1.26, 28.9)    & 9.667     & 0.002   & 0.138   & (6, 138)       & 1.893     & 0.086     & 0.012    \\ \hline
\end{tabular}
\caption{ANOVA results for all measures against \textit{ring number}, \textit{autonomy}, and their \textit{interaction}. DF = (DFn, DFd) are the degrees of freedom in the numerator and denominator respectively, p specifies the p-value, $\eta^2$ is the generalized effect size.}
\label{table:anova_results}
\end{table*}

While the demographics were included as potential covariates, preliminary analysis showed no clear effect on any dependent variable. Therefore, they were not included in the main analysis. The time of the first movement of each trial was also excluded to eliminate any adaptation effects. While we recorded pupil diameter using an eye-tracker, the data was unreliable due to device mis-calibration, and was therefore discarded from the results and analysis.

\begin{figure*}[t!]
\centering
\begin{subfigure}{0.31\textwidth}
    \includegraphics[width=\textwidth]{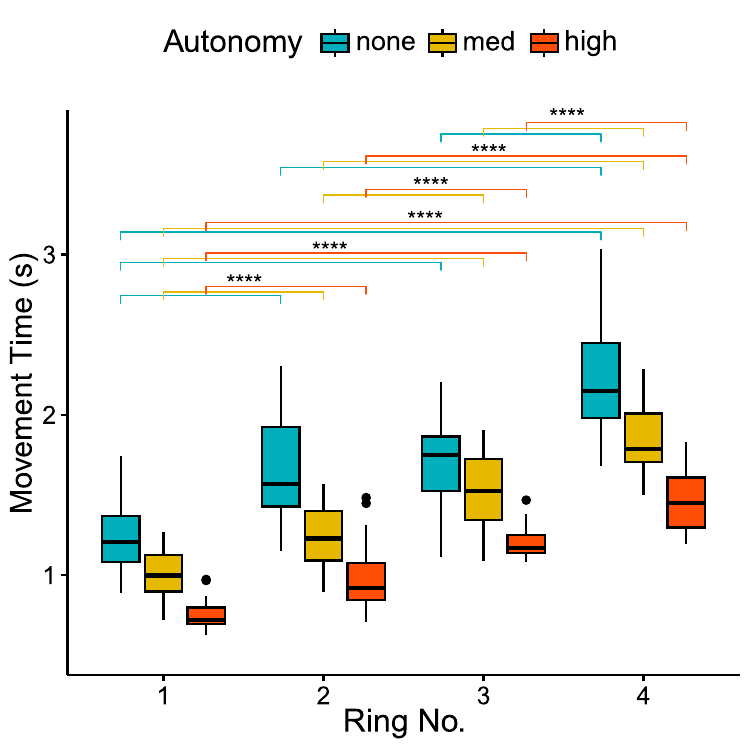}
    \caption{Movement Time}
    \label{fig:boxplot_mt}
\end{subfigure}
\begin{subfigure}{0.31\textwidth}
    \includegraphics[width=\textwidth]{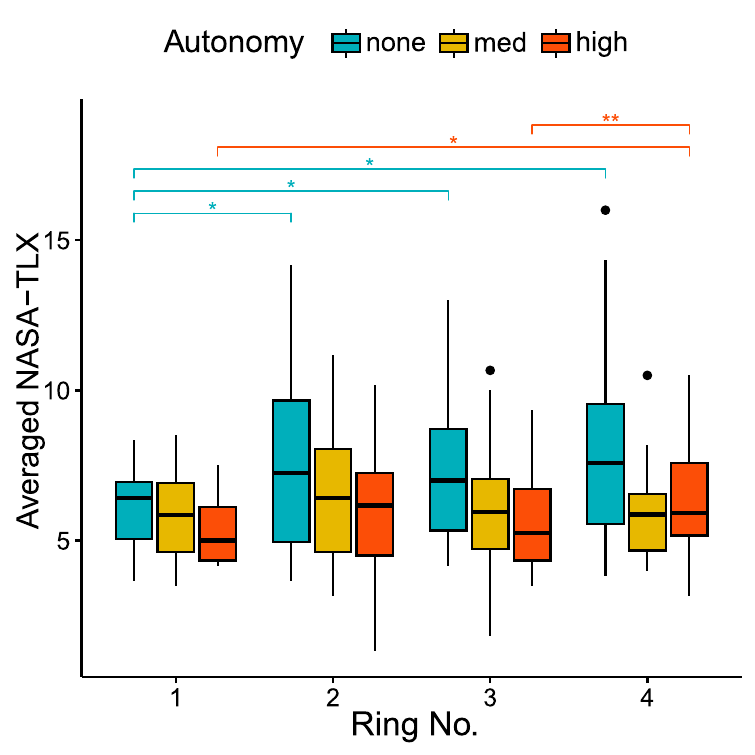}
    \caption{Averaged NASA-TLX}
    \label{fig:boxplot_tlx}
\end{subfigure}
\begin{subfigure}{0.31\textwidth}
    \includegraphics[width=\textwidth]{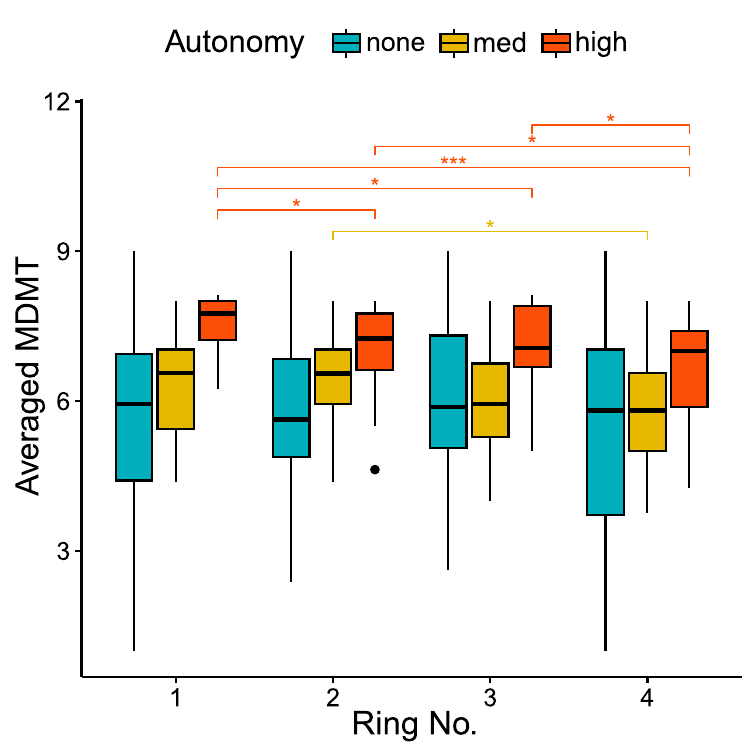}
    \caption{Averaged MDMT}
    \label{fig:boxplot_mdmt}
\end{subfigure}
\caption{Boxplots showing the distributions of each measure against \textit{ring number}, grouped by \textit{autonomy}. Differences observed from post-hoc pairwise comparisons with Holm-Bonferroni correction are also labelled with their significance levels.}
\label{fig:boxplots}
\end{figure*}

\subsection{Perceived Autonomy} \label{results:p_auto}
We evaluated if the participants' perceived autonomy was consistent with the true autonomy levels, and if this was affected by the \textit{ring number}. ANOVA indicated a clear \textit{autonomy} effect (\(F(2,46)=61.731, p<.001, \eta^2=.472\)) and a small \textit{ring number} effect (\(F(3,69)=3.530, p=.019, \eta^2=.021\)), without any interaction. Post-hoc analysis showed that the \textit{perceived autonomy} changed between all pairs of \textit{autonomy} levels across all \textit{ring numbers} (\(p<.005\)), while differences were only observed for \{\textit{1, 4}\} (\(p<.001\)) and \{\textit{2, 4}\} (\(p=.001\)) \textit{ring number} pairs under the \textit{high autonomy} level. This manipulation check suggests that participants correctly perceived changes in autonomy levels. It also suggests a trend under \textit{high autonomy} where participants' autonomy perception decreased for target rings with higher Fitts' ID.

\subsection{Movement Time} \label{results:movement_time}
ANOVA showed an effect of both \textit{ring number} (\(F(3,69)=225.503, p<.001, \eta^2=.665\)) and \textit{autonomy} (\(F(1.58,36.42)=222.807, p<.001, \eta^2=.549\)) on \textit{MT}, and an interaction effect (\(F(6,138)=8.157, p<.001, \eta^2=.061\), Figure \ref{fig:boxplot_mt}). Post-hoc analysis indicated differences in \textit{MT} for all pairs of \textit{autonomy} levels across all \textit{ring numbers} (\(p<.001\)). Clear differences were also found for all pairs of \textit{ring numbers} across all \textit{autonomy} levels (\(p<.001\)), except for the \{\textit{2, 3}\} pair of \textit{ring numbers} under \textit{no autonomy}. This suggests that Fitts' Law holds under the \textit{no autonomy} condition as the \textit{MT} increases across rings with increasing Fitts' ID (where rings \{\textit{2, 3}\} have the same Fitts' ID). However, the interaction effect shows that this increase in \textit{MT} is reduced for higher levels of \textit{autonomy}.

To gain further insight into autonomy's effect on the resulting movement, we first analyzed the recorded velocities of the user movements under different autonomy levels. ANOVA showed an effect of \textit{autonomy} on the users' \textit{average velocity} (\(F(2,46)=65.252, p<.001, \eta^2=.183\)), indicating that users moved at \textit{higher velocities} on average with increasing levels of autonomy. We also analyzed the average distance by which the robot was leading the human within each reaching trajectory, and observed an effect of \textit{autonomy} on the measure (\(F(2,46)=65.252, p<.001, \eta^2=.183\)), which is consistent with the previous result of users moving faster with more robot autonomy. 


\subsection{Cognitive Load} \label{results:cl}
We observed effects of both \textit{ring number} (\(F(3,69)=6.859, p<.001, \eta^2=.040\)) and \textit{autonomy} (\(F(1.24,28.62)=15.271, p<.001, \eta^2=.082\)) on the raw \textit{NASA-TLX} score, where the 6 subscales showed internal consistency (Cronbach's \(\alpha = .818\)). An interaction on the \textit{NASA-TLX} score was also observed (\(F(6,138)=3.823, p=.001, \eta^2=.021\), Figure \ref{fig:boxplot_tlx}). Post-hoc analysis showed differences for all pairs of \textit{autonomy} levels across all \textit{ring numbers} (\(p<.05\)), but no differences were observed with the \{\textit{no, medium}\} \textit{autonomy} level pair for \textit{ring 1} and the \{\textit{medium, high}\} \textit{autonomy} level pairs across \textit{rings} \textit{2, 3} and \textit{4}. 

These results suggest that the perceived cognitive load was lower with increased \textit{autonomy}, though this was saturated between the \textit{no} and \textit{medium} \textit{autonomy} levels for the ring with the lowest Fitts' ID, and between the \textit{medium} and \textit{high} \textit{autonomy} levels for the rings with higher Fitts' ID. Further post-hoc analysis between all pairs of \textit{ring numbers} for each \textit{autonomy} level showed some differences (Figure \ref{fig:boxplot_tlx}). Here, it is worth highlighting that under \textit{medium autonomy} participants perceived the same cognitive load across all target rings.

\subsection{Trust} \label{results:trust}
The \textit{MDMT} averaged across the \textit{Reliable} and \textit{Capable} sub-categories (Cronbach's \(\alpha=.95\)) indicated both a \textit{ring number} effect (\(F(3,69)=8.069, p<.001, \eta^2=.025\)) and an \textit{autonomy} effect (\(F(1.26,28.9)=9.667, p=.002, \eta^2=.138\)), while no interaction effect was observed (Figure \ref{fig:boxplot_mdmt}). Post-hoc analysis found differences for all pairs of \textit{autonomy} levels (\(p<.05\)) across all \textit{ring numbers}, suggesting that participants' trust in the robot was in general greater under higher \textit{autonomy}. Further post-hoc analysis showed differences under \textit{high autonomy} level across all \textit{ring number} pairs except the \{\textit{2, 3}\} pair, but no differences in general between all pairs of \textit{ring numbers} for \textit{low} and \textit{medium autonomy} levels. This indicates that while the variation in Fitts' ID did not have a clear impact on the users' trust towards the robot in general, there was a trend of users having less trust towards the \textit{high autonomy} robot for target rings with higher Fitts' ID. 



\section{Discussion} \label{discussion}

\subsection{Difficulty and Autonomy Affect Task Performance} \label{subsec:discussion1}

The results indicate that participants' \textit{movement time} increased with greater difficulty and decreased under higher robot \textit{autonomy}. Here, the results only partially support \textbf{H1}, as although the observed main effect of \textit{ring number} on \textit{movement time} confirms that Fitts' Law holds under \textit{no autonomy}, the interaction effect with \textit{autonomy} level (see Section \ref{results:movement_time}) altered this relationship such that a similar Fitts' Law trend was not observed under the influence of robot \textit{autonomy}. 


The observed impact of robot autonomy on the Fitts' relationship may have been due to several factors: i) the autonomous robot controller's target reaching motions did not satisfy Fitts' Law; and ii) the existence of robot autonomy affected the human behavior. Here, our results indicated that users moved at \textit{higher velocities} on average with increasing levels of autonomy. This suggests that under higher levels of robot autonomy users' movement was more \textit{in-sync} with the robot's higher (constant) velocity movements, which is also consistent with the observed trend of the average distance by which the robot was leading the human within each reaching trajectory decreasing across increasing autonomy levels. 
This may imply that upon perceiving the effect of robot autonomy through visual feedback at the start of each trial, the user adapted their movement behavior throughout the remaining reaching motions of the trial.
Overall, these factors may have led to the deviation of the relationship between movement time and difficulty from the original Fitts' Law prediction under the influence of robot autonomy.

\subsection{Benchmarking Shared Control Using Adapted Fitts' Law} \label{subsec:discussion2}

The effect of robot autonomy on the Fitts' relationship suggests that a more comprehensive performance model is needed to account for the impact of both Fitts' ID and robot autonomy on movement time, and to capture their interaction. While further investigation is needed to develop and refine such a model, based on our results we propose a Linear Mixed Model with both ID and \textit{robot autonomy} (\(\gamma\)) and their interaction, forming an adapted Fitts' equation:
\begin{equation}
    \mathrm{MT} = a + b_1 \mathrm{ID} + b_2 \gamma + b_3 (\gamma \cdot \mathrm{ID}),
\label{eqn:auto_fitts}
\end{equation}
where for our experiment, this model would have coefficients
\begin{align}
    [a, b_1, b_2, b_3] = [-0.28, 0.54, -0.08, -0.18]
\end{align}
with (\(R^2 = 0.735\)). 

The linear model coefficients may be interpreted in a similar manner as the original Fitts' Law. Here, $b_1$ indicates the quality of the control interface, where a smaller value indicates less increase in movement time across increasing Fitts' ID and hence a more efficient interface. $b_2$ indicates the quality of the autonomous robot algorithm, quantifying the change in task performance as more control authority is given to the algorithm. Finally, $b_3$ indicates the interaction between the control interface and the robot algorithm, where a smaller magnitude suggests a better fit and less interference between the two on task performance. In particular, a negative $b_3$ implies that higher autonomy effectively enhances the interface quality and hence the overall shared control system's efficiency, and that the performance of the robot algorithm relative to a human is better for higher difficulty levels. 

Our results clearly highlight the level of autonomy as a factor of variation in movement time. Furthermore, the equation implies that the task's \textit{effective} ID may deviate from its true Fitts' ID due to varying robot autonomy. While movement time may be directly used to quantify effective ID in the human-only case with no robot autonomy given its direct linear link to the true Fitts' ID, it is worth investigating if movement time can also be directly applied to shared control settings with varying robot autonomy, or if more complex measures accounting for autonomy are required. Moreover, the effective ID may also be linked to the user's perceived cognitive load level, which presents an opportunity for future works to determine the exact relationship between them and investigate the possibility of using effective ID as a real-time estimator of perceived cognitive load. 
Ultimately, we aim for the adapted Fitts' Law to serve as a foundation for future works where further investigation and development is needed to bring it closer to challenging real-world teleoperation tasks.

\subsection{Effects of Difficulty and Autonomy on User Perception} \label{subsec:discussion3}

Our results show that in general the subjective \textit{cognitive load} was lower under higher robot \textit{autonomy}, and the users' perceived cognitive load was higher for greater Fitts' ID, thereby partially supporting \textbf{H2}. Here, it is worth highlighting that while some differences in perceived cognitive load were observed between target rings with different Fitts' ID under both \textit{no} and \textit{high} \textit{autonomy} levels, under \textit{medium autonomy} participants perceived the same level of cognitive load across all target rings (see Section \ref{results:cl}). This is an interesting finding as it may suggest that when there is a change in task difficulty during human-robot shared control, a medium level of robot autonomy may be employed to minimize its impact on the user's perceived cognitive load. We note that this is a preliminary result and requires further investigation with different tasks and experimental scenarios to better understand this relationship.


The \textit{MDMT} results indicate that trust in the robot increased with higher \textit{autonomy} and had a trend of reducing with difficulty in the high autonomy condition. This suggests that while participants understood the robot's behavior and identified its superior performance, they may be less comfortable with it having more control authority for complex actions.

Overall, we highlight the importance of investigating the impact of different types of feedback during teleoperation on the user's ability to understand the robot's movement during task completion which may also impact their perception of the robot. Moreover, while we have considered participant expertise through the demographics items and controlled for potential learning effects through practice trials, we believe that increasing familiarity with robot operation and task completion may still affect the user's perception of the autonomous system and task difficulty, and therefore merit consideration when designing effective shared control systems.

\subsection{Generalization to Other Factors} \label{subsec:discussion4}

In this study, the robot's accuracy and efficiency in task completion may have caused participants to perceive it as more trustworthy as they are more comfortable with better task performance. This could have also led to lower levels of perceived cognitive load since the user focuses less on improving accuracy, which is consistent with previous findings in \cite{endsley1995out} that more automated systems could lead to less effort exerted by the operator. Previous works have however shown that robot error \cite{desai2012error,salem2015faulty} can indeed impact user perception and performance during the interaction. Therefore, while the movements of the user and robot were more in-sync under higher autonomy levels in this study, it is worth investigating if this trend arose from participants' trust towards accurate robot performance, and whether alternative behaviors emerge as a result of the impacted user perception due to imperfect robot behavior.

Furthermore, it is worth noting that our observed results are likely to be feedback-reliant. Here, previous works have demonstrated that providing haptic feedback to the user during shared control can enhance task performance \cite{zhang2021haptic, thomas2023haptic} and decrease cognitive workload \cite{thomas2023haptic}. \cite{kovacs2008perceptual} also showed through a similar Fitts' reaching task that performance can be impacted by the size and placement of the task's visual display. Moreover, while our setup involving high-frequency communication and tuned PID joint controllers minimized the delay between user input and the resulting robot motion, teleoperation setups with more realistic delay may also impact the task performance and user perception \cite{louca2024impact}, especially in tasks such as repeated target reaching which require high precision under time constraints. Therefore, future works should investigate how the type and quality of feedback affect task performance and user perception of the robot in shared control settings.

Finally, while the chosen target reaching task in this study enabled adjustable levels of difficulty based on Fitts' Law and allowed us to systematically evaluate the effects of robot autonomy on user performance, it is possible that our results may be task-specific, and different findings may be observed from alternative tasks and setups. Therefore, we emphasize the need for further investigation with more complex, realistic teleoperation tasks to achieve a more robust human-robot performance model.
Similarly, we note that differing results may arise from other methods of sharing control than our implementation of the blending formulation, such as by inferring human intent \cite{wang2021intent} or controlling the robot through optimization techniques \cite{selvaggio2021shared}. Therefore, it is important for future works to validate our proposed benchmarking method in other task scenarios and control-sharing mechanisms in order to better generalize our findings.

\section{Conclusion}
This paper explored the relationship between task difficulty, robot autonomy, human cognitive load and trust and the overall task performance during a HRC task. Our hypothesis of observing higher movement times and cognitive load levels in more difficult tasks was confirmed. Additionally, our analysis revealed that robot assistance improved task performance and decreased users' perceived cognitive workload, with an interaction between task difficulty and robot autonomy on performance also observed. Through proposing an adapted Fitts' Law as a benchmarking method to formally evaluate the quality of shared control systems and their impact on the user in collaborative tasks, we aim for future works to further investigate and validate this relationship in different scenarios, thereby leading to better generalization of our findings and the development of more intelligent and adaptive robots for HRC.


\bibliographystyle{ieeetr}
\bibliography{references}


\end{document}